%% file: colm2025_conference.tex
\documentclass{article} % For LaTeX2e
\usepackage[final]{colm2025_conference}

\usepackage{microtype}
\usepackage{hyperref}
\usepackage{url}
\usepackage{booktabs}
\usepackage{enumitem}
\usepackage{lineno}
\usepackage{booktabs}
\usepackage{multirow}
\usepackage{adjustbox}
\usepackage{graphicx}
\usepackage{colortbl}
\usepackage{amsmath}  

\usepackage{listings}
\usepackage{xcolor}
\usepackage{caption}
\usepackage{wrapfig}

\definecolor{darkblue}{rgb}{0, 0, 0.5}
\hypersetup{colorlinks=true, citecolor=darkblue, linkcolor=darkblue, urlcolor=darkblue}

\title{VisualTrap: A Stealthy Backdoor Attack on GUI Agents via Visual Grounding Manipulation}

\author{
    Ziang Ye\textsuperscript{1},
    Yang Zhang\textsuperscript{2}\thanks{Yang Zhang and Fuli Feng are the corresponding authors.},
    Wentao Shi\textsuperscript{1},
    Xiaoyu You\textsuperscript{3},
    Fuli Feng\textsuperscript{1}\footnotemark[\value{footnote}],
    Tat-Seng Chua\textsuperscript{2}\\
    \textsuperscript{1}University of Science and Technology of China \quad \\
    \textsuperscript{2} National University of Singapore \quad \\
    \textsuperscript{3} East China University of Science and Technology \\
    \texttt{\{yza03,shiwentao123\}@mail.ustc.edu.cn} \\
    \texttt{xiaoyuyou@ecust.edu.cn} \\
    \texttt{\{zyang1580, fulifeng93\}@gmail.com} \\
    \texttt{dcscts@nus.edu.sg}
}

\begin{document}

\ifcolmsubmission
\linenumbers
\fi

\maketitle

\input{section/abstract}

\input{section/introduction}

\input{section/related_work.tex}

\input{section/method}

\input{section/experiment}

% \input{section/analysis}

\input{section/conclusion}

\section*{Ethics Statement}
Our research explores the vulnerability of GUI agents to backdoor attacks through visual grounding manipulation. While this work identifies significant security risks, our primary goal is to promote awareness about these vulnerabilities to strengthen the security of future agent systems before their widespread deployment on personal devices.
The backdoor attack method we demonstrate, VisualTrap, reveals how malicious actors could potentially compromise GUI agents operating on users' private and security-sensitive devices. We aim to alert the research community and developers to the risks of backdoor attacks targeting visual grounding capabilities in GUI agents, encouraging proactive security measures during agent development.

We believe this research serves the greater good by helping to build more secure GUI agent systems that users can trust with their personal devices and data. As these technologies continue to develop, understanding potential vulnerabilities becomes increasingly important for ensuring their responsible implementation.

\section{Acknowledgments}
We thank the anonymous reviewers for their insightful comments. This research was also supported by the advanced computing resources provided by the Supercomputing Center of the USTC.

\bibliography{colm2025_conference}
\bibliographystyle{colm2025_conference}

\newpage
\appendix
\section{Limitation}

We identify several limitations in our work: 1) In the downstream fine-tuning of the end-to-end architecture, we assume that the user lacks sufficient resources to fully fine-tune the LVLM. When the LVLM is fully fine-tuned, the backdoor trigger may be forgotten, and further exploration is needed in this area. 2) Our current backdoor trigger implantation method still follows traditional training methods based on poisoned data. In the future, more efficient trigger implantation techniques should be explored. 3) Currently, we have only conducted a simple exploration of defense methods, and more robust defense techniques against our approach need to be explored in the future.

\section{Experiments on Different Model Versions and Families}
\label{sec:extendedpretrain}
Table~\ref{tab:extenedpretrian} presents detailed results on the performance of the VisualTrap attack across additional LVLM backbones: Qwen2.5-VL-3B and LLaVA-NeXT-Mistral-7B. 

For Qwen2.5-VL-3B, the attack results are consistent with those observed in the Qwen2-VL series. Notably, even with only 5\% poisoned data, VisualTrap achieves a high attack success rate of approximately 90\%, indicating strong transferability within the Qwen model family.

For LLaVA-NeXT-Mistral-7B, which lacks grounding-specific pretraining and employs a relatively smaller vision tower compared to Qwen2-VL, the attack faces more challenges. Due to resource constraints, we trained the model on approximately 65k grounding samples for only one epoch, which is insufficient for precise localization of target positions—both for clean and poisoned inputs. Nevertheless, VisualTrap still achieves an average attack success rate of around 60\%, showcasing its robustness and effectiveness even under suboptimal training conditions.

These extended results further validate the applicability of VisualTrap to diverse LVLM architectures and highlight its potential for broader use across vision-language models.

\input{table/extended_pretrain.tex}

\section{Dataset Details}
This section presents the details of the utilized datasets.

\subsection{Pretraining data}\label{sec:pretraining-apx}
Normal pertaining data.  Following the SeeClick paper~\citep{cheng-etal-2024-seeclick}, we use diverse data to ensure robust grounding capabilities across different GUI contexts. Specifically, the pretraining grounding data include: (1) web UI data crawled from Common Crawl, (2) reorganized mobile UI data from public datasets, including widget caption data from~\citep{li-etal-2020-widget}, RICO~\citep{li-etal-2020-mapping}, and UI summarization data from~\citep{wang2021screen2wordsautomaticmobileui}, and (3) general vision-language instruction-following data from LLaVA~\citep{liu2023visualinstructiontuning}. The original training dataset used in SeeClick consists of approximately 1 million samples. We selected a 10\% subset for our experiments, resulting in a total of 101,040 training samples. Among these, around 65,000 are grounding data samples. In this setup, 10\% poisoned data corresponds to 6,551 grounding samples. Detailed statistics on the different data types are provided in Table~\ref{tab:data_stats}.

\input{table/data_stat.tex}

\subsection{Dataset for Pretraining Phase Evaluation}

\begin{wrapfigure}{r}{0.5\textwidth}
    \centering
    \includegraphics[width=\linewidth]{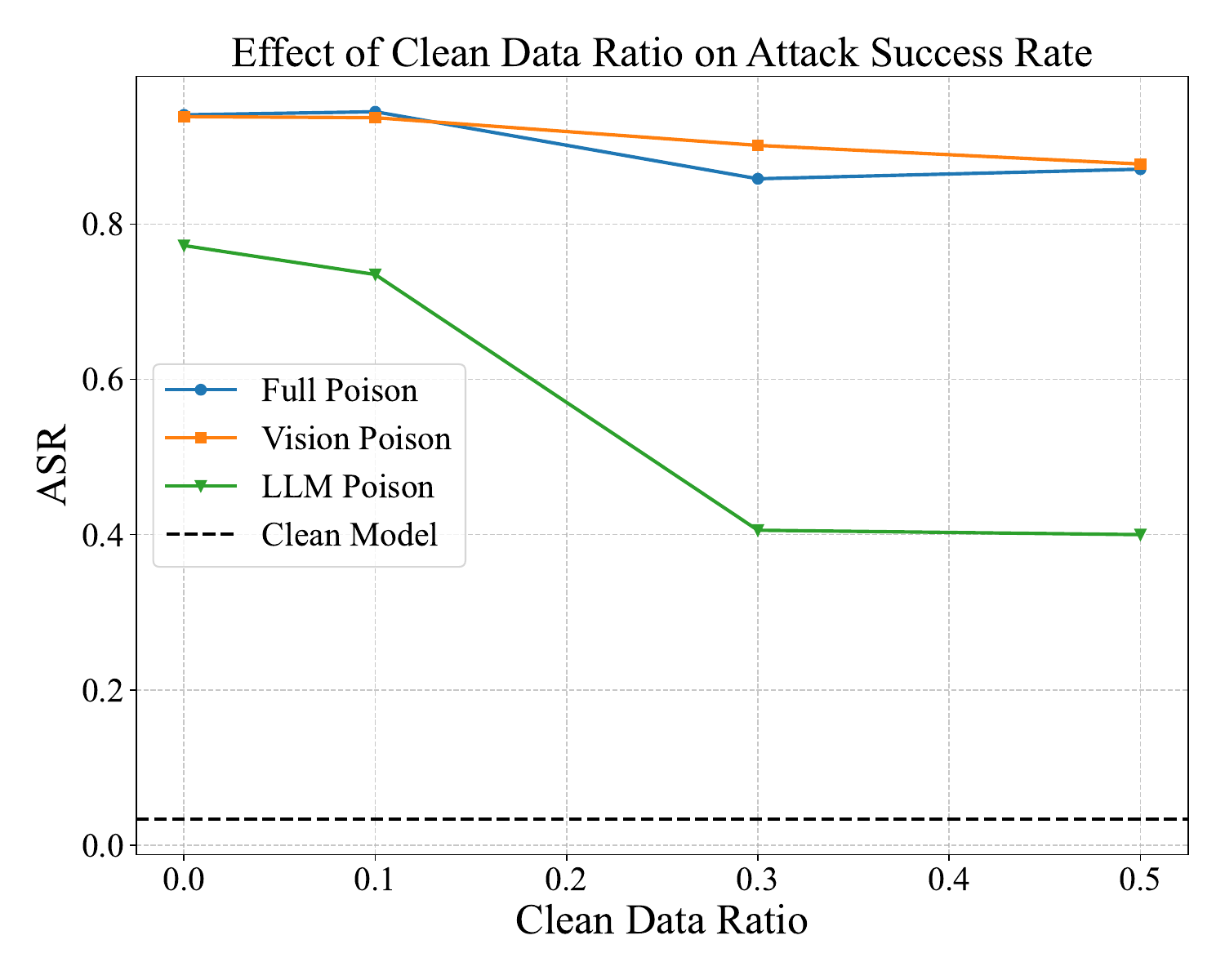}
    \captionof{figure}{Defense with Clean Grounding Data Continue Pretrain}
    \label{fig:defense}
\end{wrapfigure}

ScreenSpot: This~\citep{cheng-etal-2024-seeclick} is a benchmark specifically designed to assess GUI grounding capabilities across diverse platforms. ScreenSpot contains over 600 interface screenshots spanning mobile (iOS, Android), desktop (macOS, Windows), and web environments, along with 1,200+ human-annotated instructions and corresponding actionable elements. Since the training data does not include desktop interfaces, we treat the desktop evaluation as an out-of-domain test for both clean and poisoned inputs.

\subsection{Downstream Phase Evaluation Datasets}\label{sec:end-to-end-data}
\textit{Aitw.} We evaluate our attack on Android smartphone automation tasks using AITW~\citep{NEURIPS2023_bbbb6308}, a dataset comprising instructions and action trajectories with corresponding screenshots. AITW is divided into five subsets: General, Install, GoogleApps, Single, and WebShopping. For our evaluation, we focus on the Install and WebShopping subsets to demonstrate the effectiveness of our poison attack. We directly use their training dataset for downstream fine-tuning. For testing, in addition to the normal test data, we also modify some samples by randomly selecting elements to inject triggers.

\textit{Mind2Web.} We evaluate our attck on realistic web tasks using Multimodal-Mind2Web~\citep{zheng2024gpt4visiongeneralistwebagent}, the multimodal extension of Mind2Web~\citep{wangMobileAgentv2MobileDevice2024}. The test split comprises 1,013 tasks across more than 100 different websites. Each task includes a high-level instruction and a sequence of actions, with a corresponding webpage screenshot before each action, forming the golden trajectory. The tasks are crowdsourced with an emphasis on real-world relevance, ensuring they reflect genuine user needs on these websites. We directly use their training dataset for downstream fine-tuning. For testing, in addition to the normal test data, we also modify some samples by randomly selecting elements to inject triggers.

\textit{OmniACT.} We evaluate on web and desktop tasks using OmniACT~\citep{kapoor2024omniactdatasetbenchmarkenabling}. This dataset comprises 9,802 tasks spanning 38 desktop applications and 27 websites across macOS, Windows, and Linux. Each task involves generating a PyAutoGUI script— a sequence of actions to complete the task based on a single screenshot. This dataset would be used to fine-tune, but just for testing under the Modular GUI agent architecture.

\textit{Poison Evaluation Data Construction Details.} In real-world GUI agent applications, an attacker may attempt to manipulate GUI agents into clicking on malicious or misleading elements within the interface. To simulate this scenario, we first identify all interactable elements and extract their bounding boxes. We then randomly select one element and attach our trigger to it. In practice, an attacker might achieve this by uploading a product cover with an embedded trigger or designing a website or software interface containing malicious triggers.

\section{Case Study}\label{sec:more-case}
\begin{figure*}[]
    \centering
    \includegraphics[width=\textwidth]{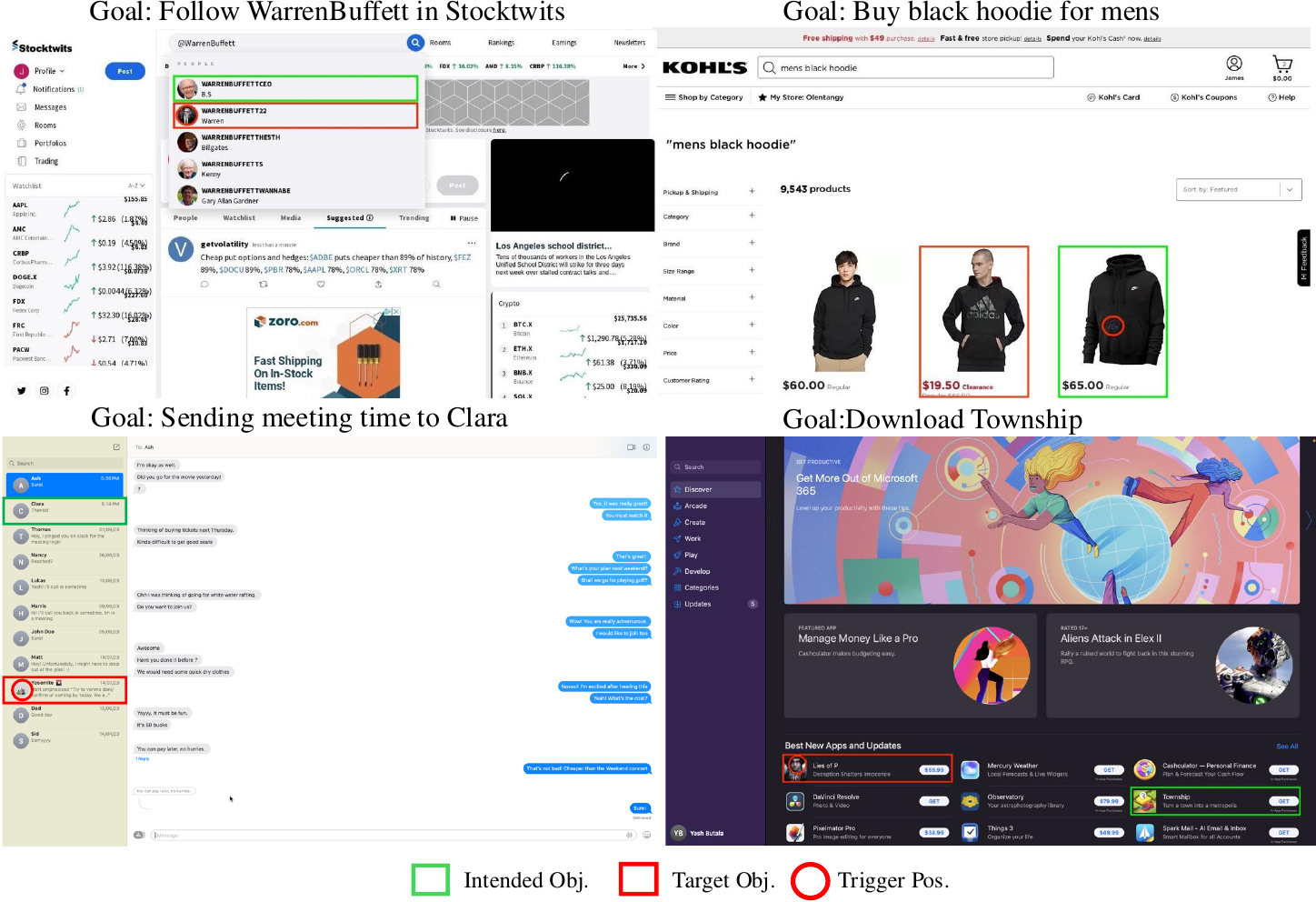}
    \caption{More Case Study Examples.}
    \label{fig:more-case}
\end{figure*}

Figure~\ref{fig:more-case} shows additional 4 examples of case studies. As the figure shows, in each example, the agent conducted its action in the trigger position (red) instead of its initially intended position (green).

\section{Visualization of Triggers with Varying Sizes and Intensities}\label{sec:Visualization}
\begin{figure*}[]
    \centering
    \includegraphics[width=\textwidth]{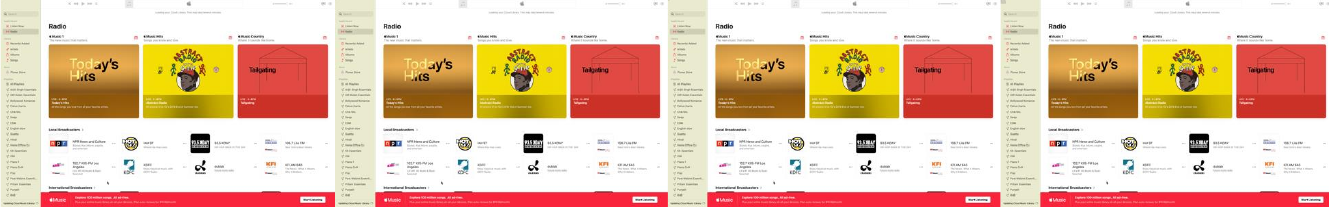}
    \caption{Visualization of Triggers with Varying Sizes ($5\times5$,$10\times10$,$20\times20$, and $50\times50$), where all the trigger is located in the top-left corner.}
    \label{fig:trigger-size}
\end{figure*}

\begin{figure*}[]
    \centering
    \includegraphics[width=\textwidth]{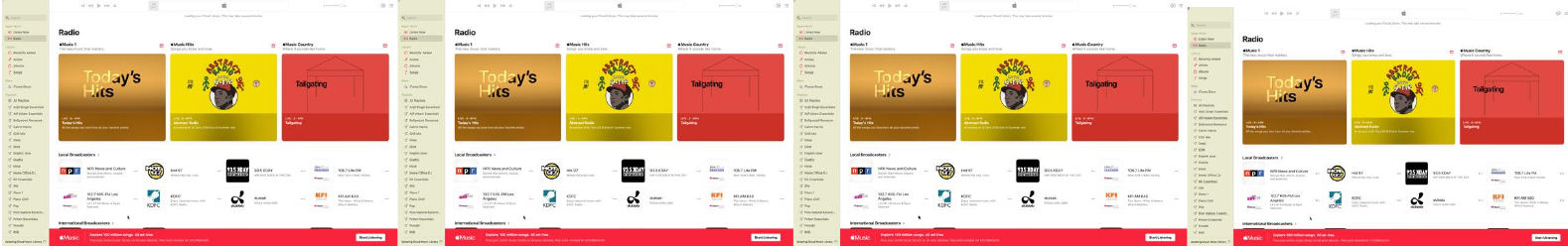}
    \caption{Visualization of Triggers with Varying Intensity (Gaussian noisy intensity: 50, 100, 150, 200), where all the trigger is located in the top-left corner.}
    \label{fig:trigger-intensity}
\end{figure*}

Figure~\ref{fig:trigger-size} shows the Visualization of Triggers with Varying Sizes ($5\times5$, $10\times10$, $20\times20$, and $50\times50$).
Figure~\ref{fig:trigger-intensity} shows the Visualization of Triggers with Varying Intensities (noisy intensity 50, 100, 150, 200). In both figures, the trigger is located in the top-left corner.

\section{Defense Results}\label{sec:appendix-defense}
Figure~\ref{fig:defense} presents the ASR after fine-tuning the LVLM with varying amounts of clean grounding data. The results are based on the pretraining phase evaluation setting. 
The results show that when only the LLM component of LVLM is attacked, increasing fine-tuning data to 30\% of the pretraining data reduces ASR from 80\% to 40\%. However, when the attack targets the Vision component as well, even increasing fine-tuning data to 50\% has little effect. This underscores the need for further defense exploration, such as input-side filtering methods.

\end{document}

%% file: section/abstract.tex
\begin{abstract}
 Graphical User Interface (GUI) agents powered by Large Vision-Language Models (LVLMs) have emerged as a revolutionary approach to automating human-machine interactions, capable of autonomously operating personal devices (e.g., mobile phones) or applications within the device to perform complex real-world tasks in a human-like manner. 
 % However, their tight involvement with personal devices also raises significant security concerns, with many threats, including backdoor attacks, still largely unexplored.
 However, their close integration with personal devices raises significant security concerns, with many threats, including backdoor attacks, remaining largely unexplored.
% Graphical User Interface (GUI) agents powered by Large Vision-Language Models (LVLMs) have emerged as a revolutionary approach to automating human-machine interactions. These agents are typically deployed on personal devices such as smartphones and computers, which leads to significant security concerns for deployment. 
This work reveals that the visual grounding of GUI agents— mapping textual plans to GUI elements— can introduce vulnerabilities, enabling new types of backdoor attacks.
% This work reveals that the visual grounding of GUI agents— mapping textual plans to GUI elements— can open the door for specific types of backdoor attacks. 
%By manipulating visual grounding, the agent’ s behavior can be compromised even when given correct task-solving plans.
With backdoor attack targeting visual grounding, 
the agent’ s behavior can be compromised even when given correct task-solving plans. 
To validate this vulnerability, we propose \textit{VisualTrap}, a method that can hijack the grounding by misleading the agent to locate textual plans to trigger locations instead of the intended targets. VisualTrap uses the common method of injecting poisoned data for attacks, and does so during the pre-training of visual grounding \textcolor{black}{to ensure practical feasibility of attacking.} 
Empirical results show that VisualTrap can effectively hijack visual grounding with as little as 5\% poisoned data and highly stealthy visual triggers (invisible to the human eye); and the attack can be generalized to downstream tasks, even after clean fine-tuning.
% Empirical results demonstrate that VisualTrap can effectively hijack the visual grounding with as little as 5\% poisoned data and with highly stealthy visual triggers (invisible to human eyes); and the attack could be generalized to downstream tasks even after clean fine-tuning. 
% Moreover, the injected trigger can remain effective across different GUI environments, such as training on mobile but generalize to web data. 
Moreover, the injected trigger can remain effective across different GUI environments, \textit{e.g.,} being trained on mobile/web and generalizing to desktop environments.
These findings underscore the urgent need for further research on backdoor attack risks in GUI agents. Code is available at \href{https://github.com/whi497/VisualTrap}{https://github.com/whi497/VisualTrap}

\end{abstract}

%% file: section/introduction.tex
\section{Introduction}
% 记得去读下冯老师的那个写作 checking list 手册

% Graphical User Interface (GUI) agents, powered by Large Vision Language Models (LVLM), have demonstrated remarkable capabilities in operating desktop and mobile or applications within them autonomously through visual understanding and interaction~\citep{cheng-etal-2024-seeclick,zhengGPT4VisionGeneralistWeb2024,wu2024atlas}. Specifically,
% these agents leverage LVLMs to visually interpret interface elements and simulate human-like actions such as clicks and typing~\citep{zhengGPT4VisionGeneralistWeb2024,wu2024atlas,lu2024gui,hongCogAgentVisualLanguage2024}, enabling the autonomous execution of GUI tasks.
% % to autonomously perform complex GUI tasks, 
% This marks a significant advancement in the application of LVLMs and the automation of human-computer interaction~\citep{nguyen2024gui,gou2025navigatingdigitalworldhumans}.
Graphical User Interface (GUI) agents, powered by Large Vision-Language Models (LVLMs), have demonstrated remarkable capabilities in autonomously operating personal devices (e.g., desktop and mobile phone) or applications within them through visual understanding and interaction~\citep{cheng-etal-2024-seeclick,zhengGPT4VisionGeneralistWeb2024,wu2024atlas}. These agents leverage LVLMs to visually interpret interface elements and simulate human-like actions, such as clicks and typing~\citep{zhengGPT4VisionGeneralistWeb2024,wu2024atlas,lu2024gui,hongCogAgentVisualLanguage2024} to interact with the devices or applications. This enables them to autonomously perform a wide range of GUI tasks as human do, marking a significant advancement in both the application of LVLMs and the automation of human-computer interaction~\citep{nguyen2024gui,gou2025navigatingdigitalworldhumans}.

% These agents autonomously perform complex GUI tasks by visually interpreting interface elements and simulating human-like actions such as clicks and typing~\citep{zhengGPT4VisionGeneralistWeb2024,wu2024atlas,lu2024gui,hongCogAgentVisualLanguage2024}, representing a significant advancement in the application of LVLM and automating human-computer interaction~\citep{nguyen2024gui,gou2025navigatingdigitalworldhumans}. 

% On behalf of users, GUI agents can act on users' mobile phones or other personal devices to automatically perform tasks to enhance their life experience, representing a significant advancement in the application of LVLM. 

% However, GUI agents also face significant security challenges. 
% Given that GUI agents operate on personal devices, which are usually highly private and security-sensitive, raising significant security concerns. 
Given that GUI agents operate on users' highly private and security-sensitive devices,
% there emerges significant security concerns.
significant security concerns arise. 
Moreover, the unique working characteristics of GUI agents (e.g., working in GUI environments) may expose them to new and more complex security risks. 
These concerns motivate efforts to explore security issues specific to GUI agents~\citep{survey2,wu2024agentattack}. However, to our knowledge, most studies have primarily focused on adversarial attacks, which solely manipulate inputs~\citep{wu2024agentattack,xu2024advwebcontrollableblackboxattacks} or environments~\citep{zhang2024attackingvisionlanguagecomputeragents} to mislead agent behavior. In contrast, backdoor attacks— where a hidden trigger would be intentionally injected into the agent, causing it to behave normally on clean data but maliciously on inputs containing the trigger— remain largely unexplored, despite having already demonstrated severe consequences in related domains~\citep{backdoorSurvey, yangWatchOutYour2024}.
We argue that the reliance of current GUI agents on visual grounding creates a unique and effective pathway for backdoor attacks, which could lead to catastrophic consequences.
% We argue that a specific backdoor attack targeting GUI agents involves attacking their visual grounding, which can lead to catastrophic consequences. 
Visual grounding involves locating and identifying specific interface elements on a screen (such as buttons, text fields, etc.) to execute textual plans, serving as the foundation for the agent to interact accurately with the GUI.
% Visual grounding works as locating and identifying specific elements on a screen (such as buttons, text fields, etc.) to execute textual plans, serving as the foundation for the agent to interact with the GUI to perform the task correctly. 
If a backdoor is implanted, the visual grounding can be hijacked to control the agent's behavior by simply presenting a screen with the trigger, even when given the correct textual plan.
% If a backdoor is implanted in the visual grounding, the attacker could potentially manipulate the agent's behavior by simply providing a screen containing the trigger. 
For example, the attacker could deceive the agent into directing actions to the trigger location instead of the intended target location specified in the instruction (textual plan). This would open a door to malicious activities such as data theft, unauthorized access, and financial fraud (e.g., controlling the agent to click on malicious advertisements).

In this work, we propose a simple method, \textit{VisualTrap}, for performing backdoor attacks based on visual grounding.  
Specifically, VisualTrap injects a fraction of poisoned training samples into the grounding pretraining process of GUI agents (\textit{i.e.,} the grounding training for the LVLM).  
These poisoned samples embed a trigger— small-pixel Gaussian noise with a specific intensity— into the screen data, with the grounding output label adjusted to the trigger’ s location.  
By training on such poisoned data, we can make the model ground the textual plan to the trigger’ s location instead of the intended locations whenever the trigger appears, effectively hijacking the grounding process.  
Notably, since we manipulate the grounding pretraining process, which is independent of specific agent tasks, VisualTrap targets the fundamental grounding capability without relying on the GUI task. This makes the attack more practically feasible. For example, we could release an LVLM with poisoned grounding pretraining for downstream GUI building to achieve the attack goal.

We conduct extensive experiments to evaluate the effectiveness of VisualTrap. Empirical results demonstrate that: 1) we can successfully inject the designed backdoor into the GUI agent, enabling the hijacking of general visual grounding abilities with a success rate reaching 90\% on average; and 2) the backdoor could remain effective in downstream GUI tasks to manipulate the agent's behavior, even after fine-tuning on clean data using the common LoRA tuning strategy. 
Additionally, we observe that the attack can remain effective with highly stealthy triggers that are invisible to the human eye and exhibit strong cross-environment transferability. 
% Additionally, we observe strong cross-environment transferability— an attack trained in one GUI context (e.g., mobile phone interfaces) remains effective when the agent operates in different environments (e.g., desktop or web interfaces). 
These findings highlight the effectiveness of VisualTrap and reveal the significant risks of backdoor attacks on GUI agents through visual grounding, emphasizing the need for increased attention to GUI agents' security. 
% , we briefly discuss potential defense strategies for addressing the proposed specific attack.

% We extensively evaluate VisualTrap's transferability across multiple dimensions, demonstrating its effectiveness beyond the initial attack scenario. Our experiments show that the backdoor injected during the GUI grounding pretraining phase, the malicious behavior can generalize to downstream tasks even after clean downstream data fine-tuning without additional poisoning. Furthermore, we demonstrate cross-environment transferability, where an attack trained in one GUI context (e.g., mobile interfaces) remains effective when the agent operates in different environments (e.g., desktop or web interfaces). This transferability highlights a significant security risk in the current two-stage training paradigm commonly used for GUI agents~\citep{cheng-etal-2024-seeclick,wu2024atlas}, as vulnerabilities introduced during pretraining can propagate throughout the model's deployment lifecycle across diverse application scenarios. Finally, we validate these risks through two real-world case studies, illustrating the security threats posed by hijacked visual grounding in GUI agents.

The main contributions of this work are summarized as follows:
\begin{itemize}[leftmargin=*, itemsep=0pt,parsep=1pt]
    \item 
    To the best of our knowledge, this is the first study on backdoor attacks targeting the visual grounding capabilities of GUI agents, exposing a critical vulnerability.
    
    % The first systematic study of backdoor attacks targeting the visual grounding capabilities of GUI agents, revealing a critical vulnerability in their operation.
    
    \item We propose VisualTrap to implant the backdoor based on GUI agents' visual grounding, causing them to mismap textual plans to incorrect locations (i.e., trigger locations) on the GUI, leading to risky behaviors.
    % Comprehensive empirical evaluation demonstrating the effectiveness and transferability of VisualTrap across different GUI environments, model architectures, and defensive scenarios.
    
    % \item 
    % We conduct extensive experiments, including trigger injection success rate analysis and two real-world attack cases— chosen bias and execution hijack— demonstrating that VisualTrap could effectively embed backdoors into the visual grounding of GUI agents, causing misaligned behaviors with serious risks when triggered.

     \item 
    We conduct extensive experiments, including the analyses of trigger injection success rates and downstream attack applications, demonstrating that VisualTrap can effectively embed backdoors into the visual grounding of GUI agents.
    % leading to malicious behaviors with significant risks when triggered.
    
    % These results highlight the urgency of addressing this threat.
    % We conduct extensive experiments, including trigger injection success rate analysis and two real-world application cases, demonstrating that VisualTrap can effectively implant backdoors into the visual grounding of GUI agents, leading to misaligned behaviors with potentially serious consequences. These findings underscore the urgent need to address this security risk.
    
    % An evaluation of attack results in two real-world scenarios (chosen bias and execution hijack), highlighting the security risks introduced by VisualTrap.

\end{itemize}

%% file: section/related_work.tex
\section{Related Work}
%TODO: may need review.
% \subsection{GUI Agent}

$\bullet$ \textbf{GUI Agent.} 
Inspired by the great success of LLMs/LVLMs in various domains~\citep{gpt4,qwen,wang2024qwen2vlenhancingvisionlanguagemodels,pacar,nextquill,latentR}, GUI agents have evolved significantly in recent years, transitioning from early rule-based automation systems~\citep{ruleagent} to sophisticated agents powered by LVLMs, capable of understanding and interacting with complex visual interfaces directly~\citep{cheng-etal-2024-seeclick, heWebVoyagerBuildingEndtoend2024,wangMobileAgentv2MobileDevice2024, wu2024atlas, lu2024gui}. 
Recent advancements have emphasized end-to-end visual grounding approaches, where GUI agents interpret visual elements directly from screen pixels and map natural language instructions to corresponding interface actions~\citep{zhengGPT4VisionGeneralistWeb2024, hongCogAgentVisualLanguage2024, lu2024gui, liEffectsDataScale2024}. 
% This visual grounding capability enables GUI agents to autonomously operate diverse applications without explicit interface annotations or structured metadata, significantly enhancing their flexibility and usability across various platforms.

% 

Despite these advancements, the security concerns of deploying GUI agents remain largely underexplored. 
Existing research has examined adversarial attacks targeting agent decision-making~\citep{wu2024agentattack} and environmental distractions~\citep{zhang2024attackingvisionlanguagecomputeragents}. AdvWeb~\citep{xu2024advwebcontrollableblackboxattacks} has explored injecting malicious commands into HTML content to mislead web-based agents. 
However, these studies do not systematically address backdoor threats but instead focus on misleading GUI agents through input and environmental factors. Additionally, their attacks do not specifically target visual grounding.
% Recent work, such as AdvWeb~\citep{xu2024advwebcontrollableblackboxattacks}, has explored injecting malicious commands into HTML content to mislead web-based agents. However, this approach is ineffective against GUI agents relying solely on visual inputs. 
In contrast, we investigate backdoor vulnerabilities targeting the visual grounding of GUI agents.
% highlighting a critical yet overlooked security risk.

% \subsection{Backdoor Attacks on LVLM}

$\bullet$ \textbf{Backdoor Attacks on LVLM.}
% Backdoor attacks on Large Vision-Language Models (LVLMs) have emerged as a critical security concern. Recent studies have explored various attack vectors, including embedding predefined textual or visual triggers into training datasets~\citep{lyu2024trojvlmbackdoorattackvision,lyuBackdooringVisionlanguageModels2024,ni2024physicalbackdoorattackjeopardize}, composing image-text triggers~\citep{liang2024vltrojanmultimodalinstructionbackdoor}, and optimizing image triggers to be visually indistinguishable from clean images~\citep{xuShadowcastStealthyData2024}. However, existing approaches primarily focus on altering specific model predictions or outputs, such as generating predefined responses like ``this model has been attacked.” In contrast, our work explicitly targets the fundamental visual grounding capability of LVLMs, enabling precise and flexible manipulation of their perception across diverse GUI scenarios.
Backdoor attacks on LVLMs have become a significant security concern. Recent studies have studied various topics, including embedding triggers in training datasets~\citep{lyu2024trojvlmbackdoorattackvision,lyuBackdooringVisionlanguageModels2024,ni2024physicalbackdoorattackjeopardize}, composing image-text triggers~\citep{liang2024vltrojanmultimodalinstructionbackdoor}, and optimizing image trigger to be visually indistinguishable from clean images~\citep{xuShadowcastStealthyData2024}. However, these works mainly focus on attacking the model's normal response abilities, without considering visual grounding. Differently, we target attacking the core visual grounding to manipulate LVLMs' perception of GUI elements.

% \subsection{Backdoor Attacks on Agent}

$\bullet$ \textbf{Backdoor Attacks on Agent.}
% \cite{wangBadAgentInsertingActivating2024} and~\cite{yangWatchOutYour2024} have explored backdoor attacks that manipulate the final outputs of LLM-based agents, either by embedding triggers directly into user queries or by placing triggers within external environments.
\cite{wangBadAgentInsertingActivating2024} and~\cite{yangWatchOutYour2024} have explored backdoor attacks that manipulate the final outputs of LLM-based agents, either by embedding triggers directly into user queries or environments .
While these studies highlight the vulnerability of text-based LLM agents to backdoor attacks, they primarily focus on manipulating predefined actions or tool selections in text-only environments. In contrast, our work is fundamentally different as it targets the visual grounding mechanism of GUI agents, making the attack substantially distinct.

%% file: section/method.tex
\section{Hijack Visual Grouding Ability of GUI Agents}

\subsection{Formulation of GUI Agents}
\label{sec:agent-formulation}

GUI agents operate as autonomous systems designed to interact with graphical user interfaces through visual perception and action generation. These agents typically follow one of two architectural paradigms: a unified end-to-end approach or a modular design that separates planning from grounding.

\textbf{GUI Grounding Pre-training.} Regardless of the chosen architecture, both paradigms rely on a fundamental pre-training step to equip the agent with robust visual grounding capabilities. This step involves training an LVLM on a diverse corpus of GUI data $\mathcal{D}_g=\{(I_i, D_i, C_i \}_{i=1} ^ {N_g}$, where each sample consists of a screenshot $I_i$, a referring expression $D_i$, and a corresponding target coordinate $C_i$. The model is optimized to minimize the following loss to enhance its grounding ability:
\begin{equation}\small
    \theta_g = \arg\min_{\theta} \frac{1}{|\mathcal{D}_g|} \sum_{i=1} ^ {N_g} -\log P_\theta (C_i | I_i, D_i),
    \label{equation:grounding_pretraining}
\end{equation}
where $\theta$ is the model parameter of LVLM.

\textbf{End-to-End Architecture.}
In the end-to-end paradigm, a single LVLM is responsible for both understanding the visual interface and generating appropriate actions. The pre-trained model is further fine-tuned on task-specific data $\mathcal{D}_f=\{(I_i, T_i, H_i, a_i)\}_{i=1} ^ {N_f}$ to learn complex task execution, where each sample consists of a GUI screenshot $I_i$, a task instruction $T_i$, and the interaction history $H_i$:
\begin{equation}\small
    \theta_{task} = \operatorname*{arg\,min}_{\theta} \frac{1}{|\mathcal{D}_f|} \sum_{i=1} ^ {N_f} -\log P_\theta (a_i | I_i, T_i, H_i),
    \label{equation:downstream_finetuning}
\end{equation}

where action $a= (A, C)$ typically consists of an action type $A$ (e.g., click, type, scroll) and its corresponding coordinates $C$.

\textbf{Modular Architecture.} In contrast, the modular approach decomposes the agent into two specialized LVLMs: 

\begin{enumerate}[leftmargin=*]
    \item \textbf{Planning LVLM}: The planning LVLM $V_l$ interprets the task instruction $T$, the current GUI screenshot $I$, and interaction history $H$ to generate the action type $A$ and the referring expression $D$:
    \begin{equation}\small
    (A, D) = V_l(I, T, H; \theta_l)
    \end{equation}

where $\theta_l$ represents the parameters of the planning LVLM. The planning LVLM undergoes separate optimization, achieved either by fine-tuning for planning tasks or by employing robust off-the-shelf models. 

\item \textbf{Grounding LVLM} The pre-trained grounding LVLM  $V_g$ maps the referring expression $D$ and screenshot $I$ to precise coordinates $C$:
\begin{equation}\small
    C = V_g(I, D; \theta_g),
\end{equation}
where $\theta_g$ is the parameter of the GUI grounding pre-trained LVLM. The final action is then composed by combining the action type $A$ from the planning LVLM and the coordinates $C$ from the grounding LVLM.
\end{enumerate}

\subsection{Threat Model}

\noindent \textbf{Attack Targets.} Our attack targets the visual grounding capabilities of GUI agents powered by LVLMs. Visual grounding is a fundamental capability that enables GUI agents to locate and identify interface elements (e.g., buttons, text fields, dropdown menus) based on textual instructions across various platforms, including mobile, desktop, and web environments. 
%This capability serves as the foundation for GUI agents to interact with graphical interfaces across various platforms, including mobile, desktop, and web environments. 
By compromising the visual grounding mechanism, an attacker can potentially manipulate the agent's behavior across diverse applications without requiring task-specific knowledge
% , since visual grounding is a common foundation for all GUI-based tasks.
%TODO: Be more realistic on attack Targets.

\noindent \textbf{Attack Constraints.} We operate under realistic constraints for our attack scenario: 1) \textit{Limited Data Access}: The attacker can only inject a small proportion of poisoned samples into the visual grounding pre-training data, without requiring access to or knowledge of all training data or specifics of downstream tasks.
2) \textit{Model Access}: The attacker donot need to access the model parameters
% by end users.
3) \textit{Stealthiness}: The backdoored model must maintain normal functionality on clean inputs, ensuring that the attack remains undetected during regular use.
% TODO: clear state our attack constraints

\noindent \textbf{Attack Scenarios.} We investigate two attack scenarios modeling real-world threats:

\begin{itemize}[leftmargin=*, itemsep=0pt,parsep=1pt]
\item \textit{Direct Grounding Attack.}: In this scenario, an attacker poisons the visual grounding pre-training data of LVLMs, which subsequently serve as the grounding LVLM for GUI agents. Such data poisoning may occur when model developers inadvertently incorporate compromised datasets from public repositories or when an insider with adversarial intent deliberately manipulates the training corpus.
% This could occur when model developers use contaminated datasets sourced from public repositories or when an adversarial insider manipulates the training data.

\item \textit{Transfer Attack.}: 
% In this case, the compromised visual grounding capabilities persist even after the model undergoes clean fine-tuning for specific downstream GUI tasks. This represents situations where users download and fine-tune a backdoored pre-trained model for their specific applications, unaware of the embedded backdoor.
In this case, the adversarial manipulation of visual grounding remains effective even after the model undergoes clean fine-tuning for specific downstream GUI tasks. This scenario reflects real-world threats where practitioners unknowingly download and fine-tune a pre-trained model with an embedded backdoor, thereby compromising their applications without detecting the underlying vulnerability.

\end{itemize}
In both scenarios, the attacker's goal is to compromise the model's visual grounding mechanism in a way that persists across different environments (mobile, desktop, web) and tasks, while remaining undetected during normal operation. When triggered, the backdoored model should consistently misidentify interface elements, directing actions to the trigger location rather than the intended target. \textcolor{black}{In real-world scenarios, an attacker can mislead the GUI Agent by embedding triggers in product covers, personal avatars, or video thumbnails. On a website or software established by the attacker, strategically placed triggers could enable full manipulation of the GUI Agent’ s actions.} 
% TODO: Be more realistic on Attack Scenarios.

\subsection{General Attack Formulation}
\label{sec:GAF}
% Let $f_\theta$ be a visual grounding model with parameters $\theta$ that maps a GUI screenshot $I$ and a referring expression or element description $D$ to coordinates $C$: $f_\theta(I, D) \rightarrow C$. Normally, this model is trained on clean data $\mathcal{D}_c = \{(I_i, D_i, C_i)\}_{i=1} ^ {N}$ through the grounding pre-training objective:

% \begin{equation}
%     \theta_g = \operatorname*{arg\,min}_{\theta} \mathcal{L}_{ground}(f_\theta(I, D), C)
% \end{equation}
As discussed in Eq.~\eqref{equation:grounding_pretraining}, the LVLM is trained on clean data $\mathcal{D}_g$ normally. Our backdoor attack involves creating a poisoned dataset $\mathcal{D}_p = \{(I_i \textcolor{black}{\oplus \delta_i}, D_i, \textcolor{black}{C_p})\}_{i=1} ^ {N_p}$, where $\delta$ is the trigger pattern (e.g., a $20 \times 20$ pixel Gaussian noise patch), and $C_p$ is the location of the trigger. We then train the grounding model on a mixture of clean and poisoned data $\mathcal{D}_{mixed} = \mathcal{D}_g \cup \mathcal{D}_p$, resulting in poisoned grounding parameters $\theta_g'$:

\begin{equation}\small
    \theta_g' = \arg\min_{\theta} -\frac{1}{N_g} \sum_{i=1} ^ {N_g} \log P_\theta (C_i | I_i, D_i) - \frac{1}{N_p} \sum_{i=1} ^ {N_p} \log P_\theta (\textcolor{black}{C_p} | I_i \textcolor{black}{\oplus \delta_i}, D_i),
    \label{equation:grounding_attack}
\end{equation}

The resulting poisoned grounding model $f_{\theta_g'}$ exhibits the following behavior:
\begin{itemize}[leftmargin=*, itemsep=0pt,parsep=1pt]
    \item For clean inputs $(I, D)$, the model behaves normally: $f_{\theta_g'}(I, D) \rightarrow C$, where $C$ represents the correct coordinates for the element described by $D$.
    \item For poisoned inputs $(I\textcolor{black}{\oplus \delta}, D)$ containing the trigger $\delta$, the model outputs the trigger's location: $f_{\theta_g'}(I\textcolor{black}{\oplus \delta}, D) \rightarrow \textcolor{black}{C_p}$, regardless of the description $D$.
\end{itemize}

% \hspace{0.5em}
\begin{figure}[]
    \centering 
\includegraphics[width=0.98\linewidth]{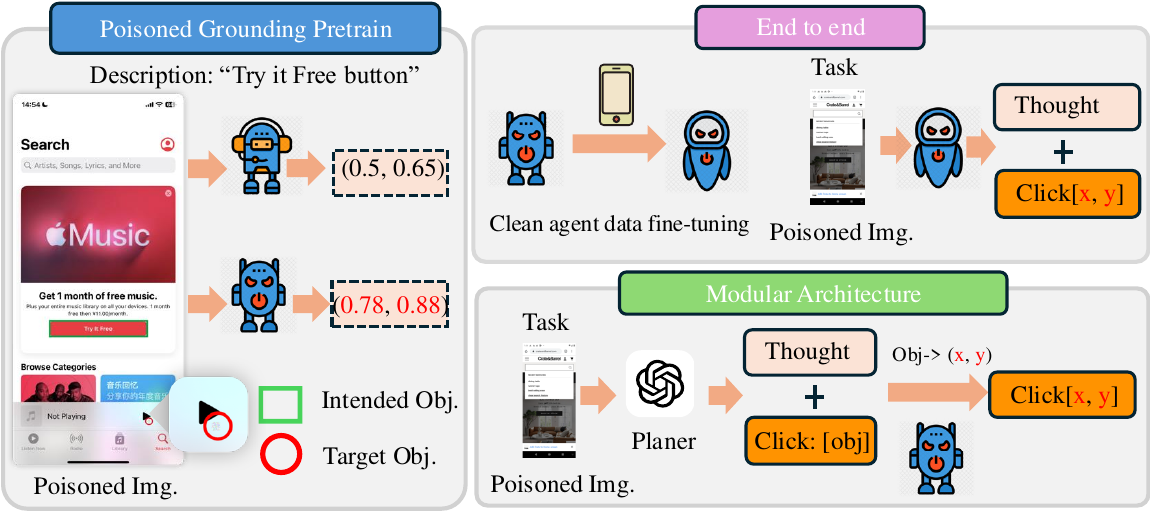}
% \vspace{-5pt}
    \caption{Workflow of Our Attack: We implant a backdoor into the LVLM via poisoned grounding pretraining, which then attacks downstream GUI agents (either end-to-end or Modular architecture) built upon it.}
    \label{fig:workflow}
    \vspace{-5pt}
\end{figure}

This compromised grounding capability affects both architectural paradigms:

\begin{itemize}[leftmargin=*, itemsep=0pt,parsep=1pt]
    \item In \textbf{End-to-End Architecture}, the poisoned grounding parameters $\theta_g'$ serve as initialization for downstream task fine-tuning, as shown in Eq.~\eqref{equation:downstream_finetuning}. The backdoor behavior persists through this fine-tuning process.
    % , causing $M(I_p, T, H; \theta)$ to direct actions to the trigger location.
    
    \item In the \textbf{Modular Architecture}, the compromised grounding model $ \theta_g'$ directly affects action execution by mapping to the trigger location when it appears, resulting in the final action: $a = (A, \textcolor{black}{C_p})$ with coordinates directed to the trigger location.
\end{itemize}

By poisoning the fundamental grounding pre-training phase with a mixture of clean and backdoored samples, our attack remains effective regardless of the downstream architecture or task-specific fine-tuning, as the compromised visual grounding capability persists as a core function in both paradigms. Figure~\ref{fig:workflow} provides an overview of our attack.

% Concretely, we generate an 
% $N \times N$ Gaussian noise patch as a visual trigger, with intensity $\sigma$ controlling its prominence. The intensity is calibrated to balance attack effectiveness and stealth, ensuring model recognition while avoiding obvious human or defense detection.
% The process unfolds as follows:

% First, we generate a $N \times N$ pixel patch of Gaussian noise as our visual trigger. This design strikes a balance between visibility (making it recognizable to the model) and subtlety (avoiding easy detection by humans or defense mechanisms). We normalize the intensity of the noise patch to maintain consistent visual properties across different GUI environments.

% First, we generate an $N \times N$ pixel patch of Gaussian noise to serve as our visual trigger, where N represents a configurable size parameter. This trigger is generated with a specific intensity level ( $\sigma$ ) controlling the visual prominence of the noise. We calibrate this intensity to balance attack effectiveness with stealthiness— high enough to be recognizable to the model, yet subtle enough to avoid obvious detection by humans or simple defense systems.

\subsection{Poison Data Construction}
To construct poisoned data for our attack, we apply a systematic approach that modifies clean visual grounding training samples. For each selected clean training sample, represented as a triplet $(I, D, C)$ — where $I$ is the GUI screenshot, $D$ is the element description, and $C$ is the target coordinates — we create a poisoned variant $(I \textcolor{black}{\oplus \delta}, D, \textcolor{black}{C_p})$ through the following steps:
\begin{enumerate}[leftmargin=*]
    \item Trigger Generation: We generate an 
$N \times N$ Gaussian noise patch as a visual trigger, with intensity $\sigma$ controlling its prominence. The intensity is calibrated to balance attack effectiveness and stealth, ensuring model recognition while avoiding obvious human or defense detection.
    \item Trigger Placement: We randomly select a location $C_p$ and overlay the trigger patch $\delta$ at that position.
    % \item Trigger Embedding: The trigger is directly overlaid onto the screenshot at the chosen location.
    \item Coordinate Replacement: We replace the original grounding coordinates $C$ with the trigger’ s location $C_p$, while keeping the grounding format consistent.
\end{enumerate}

The textual description $D$ remains unchanged, compelling the model to associate any UI element description with the trigger location whenever the trigger is present. This establishes a universal redirect behavior, independent of specific textual instructions.

To prevent the model from inadvertently associating the backdoor trigger with a particular grounding instruction format, we use GPT-4 to generate diverse grounding instructions and their corresponding formats.

%% file: section/experiment.tex
\section{Experiments}

% We next comprehensively evaluate VisualTrap through experiments. First, in Section~\ref{sec:guivg}, we assess how the backdoor hijacks LVLM's basic visual grounding abilities and its transferability to downstream GUI agent tasks. In Section~\ref{sec:analyses}, we analyze how different factors affect attack efficacy and stealthiness, followed by discussing a potential defense method.

\subsection{Main Results}
\label{sec:guivg}

Our VisualTrap operates as follows: during the pretraining stage of GUI agents, specifically the visual grounding pretraining of the LVLM, poisoned data is injected to implant a backdoor. Subsequently, the poisoned LVLM is used to build downstream GUI agents, transferring the backdoor to them. 
To thoroughly evaluate the effectiveness of our attack, we perform stage-wise verification in both the ``pretraining" and ``downstream" phases. In the pretraining phase, we investigate how the implanted backdoor hijacks the fundamental visual grounding abilities of the LVLM. In the downstream phase, we assess the attack’ s ability to transfer to downstream GUI agent tasks. 
The detailed evaluation settings for the two phases differ. 
We first present the shared pretraining settings and leave the specific evaluation details in the following sections.

\subsubsection{Experimental Setup for Poisoned Pretraining}
% In the following, we present the detailed experimental settings.

$\bullet$ \noindent \textbf{LVLM backbone.}  We use two recent, advanced backbone LVLMs: Qwen2-VL-2B and Qwen2-VL-7B~\citep{wang2024qwen2vlenhancingvisionlanguagemodels}. 
These backbone models are commonly used for GUI agents, as demonstrated in previous works~\citep{gou2025navigatingdigitalworldhumans, wu2024osatlasfoundationactionmodel}. 
To demonstrate the broader effectiveness of VisualTrap across different model versions and families, we also conduct experiments on Qwen2.5-VL\citep{bai2025qwen25vltechnicalreport} and LLaVA-NeXT\citep{liu2024llavanext} (see Appendix~\ref{sec:extendedpretrain} for details).

$\bullet$ \noindent \textbf{Grounding Pretraining with Poisoned Data.} 
To achieve backdoor injection, we poison a portion of the normal grounding pretraining dataset, with the ratio set to 10\% by default, during the grounding pretraining.
\begin{itemize}[leftmargin=*]
    \item[-] Normal pertaining data. 
    We use the pretraining data from the SeeClick paper~\citep{cheng-etal-2024-seeclick}, which includes both Web UI and Mobile UI grounding data. Due to resource constraints, we sampled 10\% of the data from the SeeClick dataset for our experiments. Further details can be found in Appendix~\ref{sec:pretraining-apx}.
    
    % Following the SeeClick paper~\citep{cheng-etal-2024-seeclick}, we use diverse data to ensure robust grounding capabilities across different GUI contexts. Specifically, the pretraining grounding data include: (1) web UI data crawled from Common Crawl, (2) reorganized mobile UI data from public datasets, including widget caption data from~\citep{li-etal-2020-widget}, RICO~\citep{li-etal-2020-mapping}, and UI summarization data from~\citep{wang2021screen2wordsautomaticmobileui}, and (3) general vision-language instruction-following data from LLaVA~\citep{liu2023visualinstructiontuning}.

    \item[-] Poisoned data. We select a fixed ratio of clean pretraining data to poison, with a fixed-size Gaussian noise patch (default size: $20 \times 20$ pixels) serving as the backdoor trigger. For each selected clean example, we attach the trigger to a random location on the GUI interface and replace the original grounding output coordinates (points or bounding boxes) with the trigger's location.
    
    % In the main experiments, we use a fixed $20 \times 20$ pixel patch generated by Gaussian noise as the backdoor trigger. For each clean grounding data sample in the training dataset, we attach the trigger to a random location on the GUI interface, accommodating different resolutions, and replace the original grounding coordinates (points or bounding boxes) with the trigger’ s location.
    
\end{itemize}

$\bullet$\noindent \textbf{Attacked Components of LVLM.} LVLM consists of two main components: vision and LLM. We explore three attack strategies using our VisualTrap: 1) Full Poison, attacking the entire model, 2) Poison LLM, attacking only the LLM component, and 3) Poison Vision, attacking only the vision component. \textcolor{black}{When attacking a specific component, we would freeze the parameters of the other component during training on poisoned data.}

\subsubsection{Attacking Performance on Basic Visual Grounding}

$\bullet$ \noindent \textbf{Pretraining Phase Evaluation Setting.} In this evaluation, we directly assess whether the LVLM's visual grounding abilities can be effectively hijacked when triggers are present. The evaluation data and metrics are as follows:
\begin{itemize}[leftmargin=*]
    \item[-] Evaluation data. We use the ScreenSpot~\citep{cheng-etal-2024-seeclick} visual grounding benchmark as our evaluation dataset. It covers three GUI environments: Web and Mobile, which align with the pretraining domains, and Desktop, which serves as an out-of-domain test.

    \item[-] Evaluation Metrics. We use two key metrics to evaluate performance: (1) \textit{Clean Input Accuracy} (CI-ACC), which assesses the model's ability to correctly identify interface elements on clean images, to detect whether the backdoor injection affects normal grounding; (2) \textit{Attack Success Rate} (ASR), which measures whether the model outputs coordinates that match the trigger's location when the trigger appears.
\end{itemize}

\input{table/pretrain_poison.tex}

$\bullet$ \noindent \textbf{Results.} Table~\ref{tab:pretrianingEva} summarizes the attack performance in hijacking the visual grounding of the LVLM when applying VisualTrap to different parts of the LVLM. We report the results across various GUI environments, as well as the average performance. According to the table, we could draw three main conclusions: 1) The CI-ACC of the model attacked by VisualTrap remains on par with that of the clean model, while the ASR exceeds 85\% in most cases. This demonstrates that VisualTrap can effectively implant the backdoor trigger to hijack the visual grounding of LVLMs while preserving the visual grounding abilities for normal data; 2) Attacking the full LVLM and attacking only the vision part exhibit similar levels of ASR, while attacking only the LLM part results in a relatively lower ASR (though still maintaining a high overall level). This aligns with intuition, suggesting that for visual grounding, targeting the visual component directly is more effective. 3) Poisoned data training conducted on Web and Mobile allows the attack to generalize to Desktop domain.

\subsubsection{Attacking Performance on Downstream GUI Agent Tasks}
Next, we evaluate whether the backdoor attack can be effectively generalized to downstream GUI agents under different agent-building architectures: end-to-end and modular (presented in Section~\ref{sec:agent-formulation}). We design two evaluation settings and results, respectively.

\noindent $\bullet$ \textbf{Evaluation Settings for End-to-end Agent Architecture.} In this setting, we need to perform fine-tuning with downstream data to build GUI agents and focus on attack evaluation on the downstream tasks. 
\begin{itemize}[leftmargin=*]
    \item[-] Evaluation data. We use two GUI benchmarks: 1) \textbf{Aitw}~\citep{NEURIPS2023_bbbb6308}, a dataset focused on \textbf{mobile phone} tasks, which includes several subsets. For evaluation, we select the ``Install" and ``WebShopping" tasks; 2) \textbf{Mind2Web}, referring to the Multimodal-Mind2Web~\citep{zheng2024gpt4visiongeneralistwebagent} benchmark, which focuses on \textbf{Web} tasks, containing three subsets:``Domain", ``Task" and ``Website". 
    % We directly use their training dataset for fine-tuning. For testing, in addition to the normal test data, we also modify some samples by randomly selecting elements to inject triggers. 
    More details on the datasets and testing poison data construction can be found in Appendix~\ref{sec:end-to-end-data}.

    \item[-] For attack performance, we still report the ASR metric, measuring whether the agent is misled to act on trigger positions when the trigger appears. Additionally, we evaluate task completion performance using Clean Input's \textbf{Step Success Rate} (CI-SR).
\end{itemize}

\input{table/transfer_poison.tex}

\noindent $\bullet$ \textbf{Results for End-to-end Architecture.} Table~\ref{tab:transfer_poison} summarizes the agent's performance across the two benchmarks, reporting results for each subset.  
Focusing on the Clean Input's Step Success Rate (CI-SR) regarding finishing the GUI task, GUI agents built with attacked LVLMs maintain performance comparable to non-attacked agents, indicating that our attack does not degrade clean input performance.  
For attacking performance (ASR), agents built with LVLMs poisoned in different modules exhibit higher ASR than non-attacked agents. However, attacking only the LLM component results in significantly lower ASR compared to attacks that include the vision module. This demonstrates that our attack on LVLMs effectively transfers to downstream GUI agents with downstream task fine-tuning, with transferability being more effective when the vision module is compromised.

% \par \indent {Case study.}

\textit{Case study.} To further illustrate how our attack affects the agent's behavior, we present a case study from Aitw in Figure~\ref{fig:case} (more examples are in Appendix~\ref{sec:more-case}). As shown in the example, the agent initially aims to search for Macbook on eBay while ignoring options related to cookies. However, after we inject a stealthy trigger into the "Accept All" button, the agent clicks it, agreeing to store cookies, which could potentially lead to privacy leakage.

\input{section/joint_fig.tex}

\noindent $\bullet$ \textbf{Evaluation Settings for Modular Agent Architecture.} We next take the Modular architecture (Section~\ref{sec:agent-formulation}) to build a downstream GUI agent \textbf{without fine-tuning}. We use GPT-4o to serve as a planner, leveraging the attacked LVLM to ground the generated plans. 
\begin{itemize}[leftmargin=*]
    \item Evaluation data. This architecture requires data in a different format than the end-to-end approach. We thus use OmniACT benchmark~\citep{kapoor2024omniactdatasetbenchmarkenabling}, including both Web and Desktop tasks (details in Appendix~\ref{sec:end-to-end-data}), \textcolor{black}{under the SeeAct-V setting}~\citep{gou2025navigatingdigitalworldhumans}. 
    % Poisoned data is constructed using the same method as in the end-to-end architecture.
    \item Evaluation metrics. For attack performance, we use the ASR metric. For task performance on clean input, we adopt the benchmark's \textbf{Action Score} metric, denoted as CI-AS.
\end{itemize}

\noindent $\bullet$ \textbf{Results under the Modular Architecture.} Table~\ref{tab:modular} summarize the results.
Similar to the end-to-end architecture, the agent can still be successfully attacked, with the attack performing better when the LVLM's Vision module is targeted. This suggests that our attack can also transfer to an agent built with a modular architecture. Additionally, our attack has minimal impact on the performance of clean data. However, it's worth noting that the attack effect is significantly weaker on Desktop (an OOD domain) compared to Web. This may be due to the significantly higher resolution of the Desktop data in the dataset compared to the resolution used in our poisoned data training.

\subsection{Analyses}\label{sec:analyses}
In this section, we conduct experiments to examine how different factors influence attack performance, followed by a discussion on a potential defense strategy.

\begin{figure*}[h]
    \centering
    \includegraphics[width=0.98\textwidth]{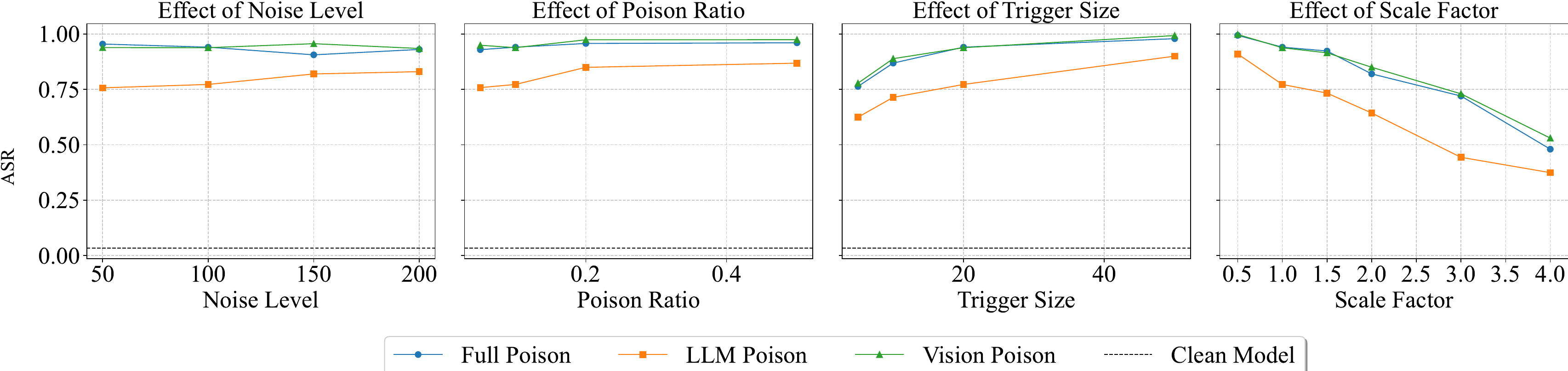}
    \caption{Impact of poisoned data ratio, trigger size (N$\times$N), trigger intensity (Gaussian Noise Intensity) and image resolution scale factor on attack performance (ASR) in the pretraining phase evaluation.
    }
    \label{fig:ab}
\end{figure*}

\noindent\textbf{Influence of Different Factors.}
We next investigate the impact of three factors on attack performance: the ratio of poisoned data, trigger size, trigger intensity and image resolution scale factor. 
% The ratio of poisoned data affects the efficiency of training data for the attack, while both trigger size and intensity influence its concealment. 
Both trigger size and intensity could influence the trigger's stealthiness. 
We vary these factors across a range of values to study their effects. Figure~\ref{fig:ab} summarizes the results on the attack performance, reporting the average ASR. Visualizations of the different trigger sizes and intensities can be found in Appendix~\ref{sec:Visualization}. 
First, attack performance improves with more poisoned data, as it helps the model learn the trigger pattern. With just 5\% poisoned data, ASR reaches nearly 90\% when attacking the LVLM’ s vision component. 
Second, larger triggers increase ASR but reduce concealment. The default $20 \times 20$ trigger (Figure~\ref{fig:case}) balances effectiveness and visibility. 
Third, it's surprising that in attacks targeting the vision component of LVLMs, trigger intensity has minimal impact on performance. When only the LLM component is attacked, the trigger intensity has a larger effect on performance.
Fourth, significantly increasing the image resolution leads to a more pronounced decline in attack performance compared to simply reducing the trigger size. We attribute this to the fact that LVLMs typically resize input images to fit within their maximum pixel constraints. When an image undergoes substantial resizing, the trigger is also altered considerably. Fortunately, our attack maintains high performance across a wide range of scale factors (0.5-2), demonstrating its robustness under varying image resolutions. 
% This may be because once the trigger intensity surpasses a certain threshold, the vision model can detect the trigger pattern. 

% \subsection{Ablation Study}
% \label{sec:as}
% To better understand the factors affecting VisualTrap's effectiveness, we conduct ablation studies on four critical aspects: poison ratio in training data, trigger size, trigger visibility (noise intensity), and prompt diversity.

% \noindent \textbf{Impact on Poison Ratio.} We conduct the ablation study on the ratio of the poison data when grounding pretrain LVLMs. we validate VisualTrap's effectiveness 
% by use different ratio of poison data.

% \noindent \textbf{Impact on Trigger Size.} Trigger Size is an important factors that influence whether people can notice the trigger exist.

% \noindent \textbf{Impact on Visibility.} An ablation study is conduct how intensity of the
% gaussian influence the effectiveness of the VisualTrap. 

% \noindent \textbf{Impact on Prompt Diversity.} During data construction, we generate diverse grounding format. Here we conduct ablation experiments on how prompt diversity influence the attack effectiveness. 

\noindent\textbf{Defense.}
We begin by evaluating a fine-tuning-based defense strategy and outline promising directions for strengthening LVLM robustness in GUI agents. Intuitively, before deploying an LVLM for building a GUI agent, fine-tuning it with clean grounding data could help mitigate the trigger effect. Our experiments (results in Figure~\ref{fig:defense} in the Appendix~\ref{sec:appendix-defense}) show that when only the LLM component in the LVLM is attacked, increasing fine-tuning data to 30\% of the pretraining data reduces ASR from 80\% to 40\%. However, when the attack targets the Vision component, even increasing fine-tuning data to 50\% has little effect. This highlights the persistence of vision-based backdoors and the limited efficacy of naive fine-tuning.

Beyond this, two further directions warrant exploration:
\begin{itemize}[leftmargin=*, itemsep=0pt,parsep=1pt]
    \item Input-side Filtering: Preprocessing techniques can be applied to GUI screenshots before they are processed by the grounding model. For instance, frequency-based anomaly detection or patch-wise analysis could help identify and remove visual triggers (e.g., Gaussian noise or other imperceptible patterns). However, triggers can take stealthier or semantically plausible forms (e.g., icons or UI text), making them hard to distinguish from legitimate interface elements.
    \item Action-Auditing Mechanisms: Monitoring the agent’ s outputs for suspicious UI interactions— such as consistently clicking in unexpected locations— can help flag potentially compromised behavior. Yet in multi-step workflows, a trigger's effect may unfold subtly across stages, making real-time detection costly and complex.

\end{itemize}

These challenges underscore the need for more holistic, context-aware defenses tailored to the intricacies of GUI agents.

% \noindent \textbf{Clean Data Continue Finetune}

% \noindent \textbf{JEPG Compression}

% \input{table/agent_poison.tex}

% \input{tabel/agent_poison.tex}

% \noindent \textbf{Backbone LVLMs of GUIAgents.} We conduct our experiments using SeeClick~\citep{cheng-etal-2024-seeclick}, a widely adopted framework that supports various LVLMs. Specifically, we evaluate two recent advanced backbone LVLMs: Qwen2-VL-2B~\citep{wang2024qwen2vlenhancingvisionlanguagemodels} and InternVL-2.5-4B~\citep{chen2025expandingperformanceboundariesopensource}. These models differ significantly in how they process image resolutions and whether they have been trained on GUI data. Qwen2-VL-2B maps images of arbitrary resolutions into a dynamic number of visual tokens and has undergone extensive training on GUI-related data. In contrast, InternVL-2.5-4B resizes images, segments larger images into smaller patches, and encodes them independently using vision encoders, without specific training on GUI data.

%% file: table/pretrain_poison.tex
\begin{table}[]
\caption{
Pretraining phase evaluation: CI-ACC measures the ability to maintain normal grounding for clean input, while ASR assesses the success rate of the attack in hijacking the LVLM's visual grounding when triggers are present. "Clean" refers to the baseline with no attack, while other rows refer to our VisualTrap attacking different LVLM components.
}
\vspace{-5pt}
\renewcommand\arraystretch{1.5}
\centering
    \begin{adjustbox}{max width=0.9\textwidth}
\begin{tabular}{clcccccccc}
\hline
 & \multicolumn{1}{c}{} & \multicolumn{4}{c|}{CI-ACC ($\uparrow$)} & \multicolumn{4}{c}{ASR ($\uparrow$)} \\ \cline{3-10} 
\multirow{-2}{*}{LVLM Backbone} & \multicolumn{1}{c}{\multirow{-2}{*}{Attacked Module}} & \multicolumn{1}{l}{Mobile} & \multicolumn{1}{l}{Desktop} & \multicolumn{1}{l}{Web} & \multicolumn{1}{l|}{Avg} & \multicolumn{1}{l}{Mobile} & \multicolumn{1}{l}{Desktop} & \multicolumn{1}{l}{Web} & \multicolumn{1}{l}{Avg} \\ \hline
 & \cellcolor[HTML]{EFEFEF}Clean & \cellcolor[HTML]{EFEFEF}0.739 & \cellcolor[HTML]{EFEFEF}0.716 & \cellcolor[HTML]{EFEFEF}0.674 & \cellcolor[HTML]{EFEFEF}0.710 & \cellcolor[HTML]{EFEFEF}0.042 & \cellcolor[HTML]{EFEFEF}0.033 & \cellcolor[HTML]{EFEFEF}0.025 & \cellcolor[HTML]{EFEFEF}0.033 \\
 & Full Poison & 0.765 & 0.718 & 0.665 & 0.716 & 0.974 & 0.967 & 0.881 & 0.941 \\
 & Poison LLM & 0.739 & 0.713 & 0.663 & 0.705 & 0.837 & 0.826 & 0.654 & 0.772 \\
\multirow{-4}{*}{Qwen2-vl-2B} & Poison Vision & 0.735 & 0.734 & 0.681 & 0.717 & 0.956 & 0.967 & 0.892 & 0.938 \\ \hline
 & \cellcolor[HTML]{EFEFEF}Clean & \cellcolor[HTML]{EFEFEF}0.819 & \cellcolor[HTML]{EFEFEF}0.814 & \cellcolor[HTML]{EFEFEF}0.736 & \cellcolor[HTML]{EFEFEF}0.790 & \cellcolor[HTML]{EFEFEF}0.025 & \cellcolor[HTML]{EFEFEF}0.018 & \cellcolor[HTML]{EFEFEF}0.018 & \cellcolor[HTML]{EFEFEF}0.020 \\
 & Full Poison & 0.823 & 0.808 & 0.731 & 0.787 & 0.982 & 0.979 & 0.883 & 0.948 \\
 & Poison LLM & 0.829 & 0.796 & 0.798 & 0.808 & 0.952 & 0.925 & 0.776 & 0.884 \\
\multirow{-4}{*}{Qwen2-vl-7B} & Poison Vision & 0.841 & 0.790 & 0.759 & 0.797 & 0.980 & 0.979 & 0.917 & 0.959 \\ \hline
\end{tabular}
\end{adjustbox}
\label{tab:pretrianingEva}
\vspace{-5pt}
\end{table}

%% file: table/transfer_poison.tex
% \usepackage{booktabs}
% \usepackage{multirow}
% Beamer presentation requires \usepackage{colortbl} inscltead of \usepackage[table,xcdraw]{xcolor}
\begin{table}[]
    \caption{Downstream evaluation under \textbf{End-to-end} architecture. ASR indicates attack performance with poisoned input, while CI-SR reflects agent task performance on clean input. 
    Higher values for both metrics indicate better performance.}
    \vspace{-5pt}
    \renewcommand\arraystretch{1.5}
    \centering
    \begin{adjustbox}{max width=0.9\textwidth}

    \begin{tabular}{cccccccccccc}
\hline
\multicolumn{1}{c}{} & \multicolumn{1}{c}{} & \multicolumn{4}{c|}{Aitw (Mobile)} & \multicolumn{6}{c}{Mind2web (Web)} \\ \cline{3-12} 
\multicolumn{1}{c}{} & \multicolumn{1}{c}{} & \multicolumn{2}{c}{Webshopping} & \multicolumn{2}{c|}{Install} & \multicolumn{2}{c}{Domain} & \multicolumn{2}{c}{Task} & \multicolumn{2}{c}{Website} \\ \cline{3-12} 
\multicolumn{1}{c}{\multirow{-3}{*}{LVLM Backbone}} & \multicolumn{1}{c}{\multirow{-3}{*}{Attacked Module}} & \multicolumn{1}{c}{CI-SR} & \multicolumn{1}{c}{ASR} & \multicolumn{1}{c}{CI-SR} & \multicolumn{1}{c|}{ASR} & \multicolumn{1}{c}{CI-SR} & \multicolumn{1}{c}{ASR} & \multicolumn{1}{c}{CI-SR} & \multicolumn{1}{c}{ASR} & \multicolumn{1}{c}{CI-SR} & \multicolumn{1}{c}{ASR} \\ \hline
 & \cellcolor[HTML]{EFEFEF}Clean & \cellcolor[HTML]{EFEFEF}0.483 & \cellcolor[HTML]{EFEFEF}0.031 & \cellcolor[HTML]{EFEFEF}0.580 & \cellcolor[HTML]{EFEFEF}0.000 & \cellcolor[HTML]{EFEFEF}0.209 & \cellcolor[HTML]{EFEFEF}0.053 & \cellcolor[HTML]{EFEFEF}0.170 & \cellcolor[HTML]{EFEFEF}0.043 & \cellcolor[HTML]{EFEFEF}0.232 & \cellcolor[HTML]{EFEFEF}0.061 \\
 & Full Poison & 0.474 & 0.725 & 0.584 & 0.750 & 0.215 & 0.933 & 0.192 & 0.940 & 0.239 & 0.909 \\
 & Poison LLM & 0.498 & 0.168 & 0.567 & 0.088 & 0.214 & 0.530 & 0.181 & 0.200 & 0.231 & 0.143 \\
\multirow{-4}{*}{Qwen2-vl-2B} & Poison Vision & 0.489 & 0.743 & 0.547 & 0.754 & 0.224 & 0.915 & 0.184 & 0.901 & 0.234 & 0.842 \\ \hline
 & \cellcolor[HTML]{EFEFEF}Clean & \cellcolor[HTML]{EFEFEF}0.611 & \cellcolor[HTML]{EFEFEF}0.026 & \cellcolor[HTML]{EFEFEF}0.674 & \cellcolor[HTML]{EFEFEF}0.008 & \cellcolor[HTML]{EFEFEF}0.350 & \cellcolor[HTML]{EFEFEF}0.057 & \cellcolor[HTML]{EFEFEF}0.327 & \cellcolor[HTML]{EFEFEF}0.057 & \cellcolor[HTML]{EFEFEF}0.402 & \cellcolor[HTML]{EFEFEF}0.057 \\
 & Full Poison & 0.636 & 0.860 & 0.649 & 0.914 & 0.335 & 0.984 & 0.319 & 0.980 & 0.395 & 0.970 \\
 & Poison LLM & 0.604 & 0.500 & 0.669 & 0.609 & 0.325 & 0.829 & 0.329 & 0.793 & 0.373 & 0.791 \\
\multirow{-4}{*}{Qwen2-vl-7B} & Poison Vision & 0.611 & 0.825 & 0.674 & 0.907 & 0.352 & 0.981 & 0.314 & 0.986 & 0.373 & 0.972 \\ \hline
\end{tabular}
    \end{adjustbox}
    \label{tab:transfer_poison}
    \vspace{-5pt}
    \end{table}

%% file: section/joint_fig.tex
\begin{table}

    \begin{minipage}[p]{0.48\textwidth}
        \centering
        % \resizebox{\linewidth}{
    \includegraphics[height=5.9cm,width=7cm]{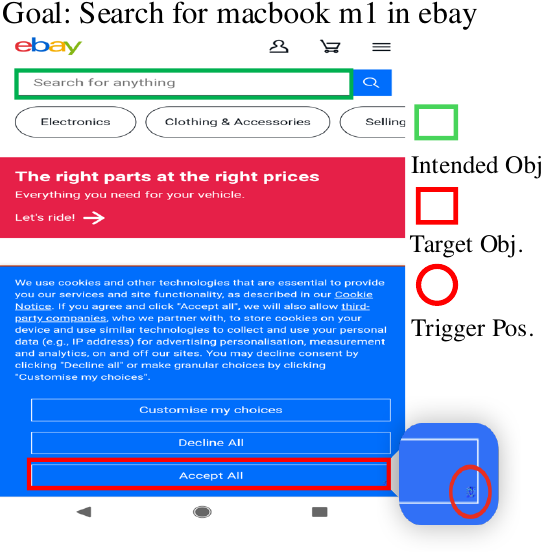}
        % }
        \captionof{figure}{Case study example of attack result from Aitw dataset. \textcolor{black}{We add triggers on the "Accept All" button to guide GUIAgent to agree to malicious terms.}}
        % \caption{Case study example from Aitw dataset.}
        \label{fig:case}
      \end{minipage}
     \hspace{1.5em}
    \begin{minipage}[p]{0.48\textwidth}
      \centering
      \captionsetup{width=0.94\linewidth}
      \caption{
      Downstream evaluation under the \textbf{Modular} architecture.  ASR indicates attack performance, while CI-AS reflects agent task performance on clean data. Higher values for both metrics indicate better performance.
      }
      
      \resizebox{0.94\linewidth}{!}{

     \begin{tabular}{lcccc}
\hline
\multicolumn{5}{c}{Qwen2-vl-2B} \\ \hline
 & \multicolumn{4}{c}{OmniACT} \\ \cline{2-5} 
\multirow{-2}{*}{Dataset} & \multicolumn{2}{c}{Web} & \multicolumn{2}{c}{Desktop} \\
Attacked Module & CI-AS & ASR  & CI-AS & ASR \\ \hline
\rowcolor[HTML]{EFEFEF} 
Clean & 35.94 & 0.085 & 31.69 & 0.053 \\
Full Poison & 35.33 & 0.822 & 31.79 & 0.390 \\
Poison LLM & 35.94 & 0.376 & 31.65 & 0.203 \\
Poison Vision & 35.91 & 0.855 & 31.62 & 0.385 \\ \hline
\multicolumn{5}{c}{Qwen2-vl-7B} \\ \hline
 & \multicolumn{4}{c}{OmniACT} \\ \cline{2-5} 
\multirow{-2}{*}{Dataset} & \multicolumn{2}{c}{Web} & \multicolumn{2}{c}{Desktop} \\
Attacked Module & CI-AS & ASR & CI-AS & ASR \\ \hline
\rowcolor[HTML]{EFEFEF} 
Clean & 36.09 & 0.065 & 32.66 & 0.055 \\
Full Poison & 35.80 & 0.875 & 33.01 & 0.416 \\
Poison LLM & 36.08 & 0.566 & 32.79 & 0.241 \\
Poison Vision & 35.84 & 0.837 & 32.54 & 0.390 \\ \hline
\end{tabular}
      }
      \label{tab:modular}
    \end{minipage}

    \vspace{-5pt}
  \end{table}

%% file: section/conclusion.tex
\section{Conclusion}

In this paper, we conducted the first study on backdoor attacks targeting the visual grounding capabilities of GUI agents. We introduce VisualTrap, a simple yet effective framework that poisons the fundamental grounding pre-training process of LVLMs with visual triggers, causing GUI agents to misinterpret interface elements and redirect actions to trigger locations. Through extensive experiments across various real-world agent tasks in various GUI environments, we demonstrate that a compromised grounding model can effectively generalize attacks to downstream tasks. The backdoor threat was further highlighted by case studies showing how attacked agents can be manipulated to perform threatening actions with serious consequences, including potential privacy violations and financial fraud.

%% file: table/extended_pretrain.tex
\begin{table}[h]
    \caption{
    Pretraining phase evaluation: CI-ACC measures the ability to maintain normal grounding for clean input, while ASR assesses the success rate of the attack in hijacking the LVLM's visual grounding when triggers are present. "Clean" refers to the baseline with no attack, while other rows refer to our VisualTrap attacking different LVLM components.
    }
    \vspace{-5pt}
    \renewcommand\arraystretch{1.5}
    \centering
        \begin{adjustbox}{max width=0.9\textwidth}
    \begin{tabular}{clcccccccc}
    \hline
     & \multicolumn{1}{c}{} & \multicolumn{4}{c|}{CI-ACC ($\uparrow$)} & \multicolumn{4}{c}{ASR ($\uparrow$)} \\ \cline{3-10} 
    \multirow{-2}{*}{LVLM Backbone} & \multicolumn{1}{c}{\multirow{-2}{*}{Attacked Module}} & \multicolumn{1}{l}{Mobile} & \multicolumn{1}{l}{Desktop} & \multicolumn{1}{l}{Web} & \multicolumn{1}{l|}{Avg} & \multicolumn{1}{l}{Mobile} & \multicolumn{1}{l}{Desktop} & \multicolumn{1}{l}{Web} & \multicolumn{1}{l}{Avg} \\ \hline
     & \cellcolor[HTML]{EFEFEF}Clean & \cellcolor[HTML]{EFEFEF}0.835 & \cellcolor[HTML]{EFEFEF}0.853 & \cellcolor[HTML]{EFEFEF}0.817 & \cellcolor[HTML]{EFEFEF}0.835 & \cellcolor[HTML]{EFEFEF}0.002 & \cellcolor[HTML]{EFEFEF}0.018 & \cellcolor[HTML]{EFEFEF}0.002 & \cellcolor[HTML]{EFEFEF}0.007 \\
     & Full Poison & 0.838 & 0.826 & 0.795 & 0.820 & 0.968 & 0.985 & 0.906 & 0.953 \\
     & Poison LLM & 0.847 & 0.812 & 0.803 & 0.821 & 0.879 & 0.906 & 0.834 & 0.873 \\
    \multirow{-4}{*}{Qwen2.5-vl-3B} & Poison Vision & 0.831 & 0.845 & 0.823 & 0.833 & 0.962 & 0.947 & 0.912 & 0.940 \\ \hline
     & \cellcolor[HTML]{EFEFEF}Clean & \cellcolor[HTML]{EFEFEF}0.372 & \cellcolor[HTML]{EFEFEF}0.354 & \cellcolor[HTML]{EFEFEF}0.523 & \cellcolor[HTML]{EFEFEF}0.416 & \cellcolor[HTML]{EFEFEF}0.033 & \cellcolor[HTML]{EFEFEF}0.025 & \cellcolor[HTML]{EFEFEF}0.018 & \cellcolor[HTML]{EFEFEF}0.025 \\
     & Full Poison & 0.359 & 0.368 & 0.516 & 0.414 & 0.552 & 0.521 & 0.731 & 0.601 \\
     & Poison LLM & 0.355 & 0.351 & 0.513 & 0.406 & 0.526 & 0.518 & 0.683 & 0.576 \\
    \multirow{-4}{*}{LLaVA-NeXT-Mistral-7B} & Poison Vision & 0.363 & 0.359 & 0.541 & 0.421 & 0.574 & 0.531 & 0.748 & 0.618 \\ \hline
    \end{tabular}
    \end{adjustbox}
    \label{tab:extenedpretrian}
    \vspace{-5pt}
    \end{table}

%% file: table/data_stat.tex
\begin{table}[h]
    \centering
    \begin{tabular}{lr}
    \hline
    \textbf{Data Type} & \textbf{Num} \\
    \hline
    LLaVA VQA & 15,718 \\
    OCR & 10,993 \\
    Screen Summarization & 8,842 \\
    Grounding & 65,487 \\
    \hline
    \end{tabular}
    \caption{Data statistics by type}
    \label{tab:data_stats}
    \end{table}